\documentclass[10pt, letterpaper]{article}
\pdfoutput=1
\usepackage[margin=1in]{geometry}
\usepackage[dvipsnames]{xcolor}
\usepackage{amsmath,amssymb,amsthm}
\usepackage{amsfonts}
\usepackage{asymptote}
\usepackage{cancel}
\usepackage{enumitem}
\usepackage{footnotebackref}
\usepackage{graphicx}
\usepackage{hyperref}
\usepackage{mathdots}
\usepackage{pgffor}
\usepackage{subfiles}
\usepackage{tikz-cd}
\usepackage{todonotes}
\usepackage{xr}
\usepackage{esint}
\usepackage{amsmath,amsfonts,amssymb,amsthm}
\usepackage{mathtools}
\usepackage{braket}
\usepackage{hyperref}
\usepackage{wrapfig}
\usepackage{enumitem}
\usepackage{tocloft}
\usepackage{microtype}
\usepackage{stmaryrd}
\usepackage[utf8]{inputenc}
\usepackage{physics}
\usepackage{graphicx}
\usepackage{setspace}
\usepackage{changepage}
\usepackage{hyphenat}
\usepackage{caption}
\usepackage{float}
\usepackage[linesnumbered,ruled,vlined]{algorithm2e}
\usepackage[noend]{algpseudocode}
\usepackage{setspace}
\usepackage{etoolbox}
\usepackage{tcolorbox}
\usepackage{tikz,tikz-qtree}
\usepackage{stackengine}
\usepackage{bbm}
\usepackage{csquotes}
\usepackage{subcaption}
\stackMath

\usetikzlibrary{shapes.misc,shapes.geometric,arrows,positioning,calc,backgrounds}

\tikzstyle{bubble} = [rectangle, rounded corners, minimum width=2cm, minimum height=1cm, text centered, draw=black, fill=black!5!white, text width=2cm]
\tikzstyle{arrow} = [thick, ->, >=stealth]
\tikzstyle{gateset} = [rectangle, minimum width=2cm, text centered, minimum height=1cm, fill=white, text width=2cm]

\SetAlFnt{\small}
\SetAlCapFnt{\normalsize}
\SetAlCapNameFnt{\normalsize}

\makeatletter
\patchcmd{\@algocf@start}% <cmd>
  {-1.5em}% <search>
  {0pt}% <replace>
  {}{}% <success><failure>
\makeatother

\DontPrintSemicolon

\SetKwComment{Comment}{\color{green!50!black}// }{}

\SetKwProg{Function}{function}{}{}

\hypersetup{
    pdftitle={Automatic Curriculum Learning Report},
    pdfauthor={Ryan Campbell \& Junsang Yoon},
    colorlinks=true,
    linkcolor=blue!80!black,
    filecolor=blue!80!black,      
    urlcolor=blue!80!black,
    citecolor=blue!80!black
}

\newtheoremstyle{thrmstyle}
  {0pt} % Space above
  {20pt} % Space below
  {\em} % Body font
  {} % Indent amount
  {\bfseries} % Theorem head font
  {.} % Punctuation after theorem head
  {.5em} % Space after theorem head
  {} % Theorem head spec
\theoremstyle{thrmstyle}

\usepackage{listings,color,textcomp}

\tcbuselibrary{most}
\tcbset{colback=black!3!white,colframe=black, enhanced, boxrule=0.2mm, arc=0.75mm}

\begin{document}

\author{
Ryan Campbell\footnote{Contributions from Ryan Campbell: wrote all of the code; ran all of the experiments; generated all of the visuals and figures; formalized the teacher-student ACL algorithm; thought of idea of using gradient reward signals; writeup}\\[.5ex]
{\normalsize University of California, Berkeley}\\
{\normalsize \texttt{ryancampbell@berkeley.edu}}
\and
\qquad
\and
Junsang Yoon\footnote{Contributions from Junsang Yoon: Project direction and framework, idea generation. Analysis for experiments, writeup. }\\[.5ex]
{\normalsize University of California, Berkeley}\\
{\normalsize \texttt{junyoon@berkeley.edu}}
}

% \author{
% Ryan Campbell\\[.5ex]
% {\normalsize University of California, Berkeley}\\
% {\normalsize \texttt{ryancampbell@berkeley.edu}}
% }

\date{}

\title{\vspace{-20pt}\bfseries{\Large Automatic Curriculum Learning with Gradient Reward Signals}}

\maketitle

\begin{abstract}

% \noindent This paper investigates the impact of using gradient norm reward signals in the context of Automatic Curriculum Learning (ACL) for deep reinforcement learning (DRL). We introduce a novel framework where the `teacher' model, utilizing the gradient norm information of a `student' model, dynamically adapts the learning curriculum. This approach is based on the hypothesis that gradient norms can provide a nuanced and effective measure of learning progress. Our experimental setup involves several reinforcement learning environments (PointMaze, AntMaze, and AdroitHandRelocate), to assess the efficacy of our method. We analyze how gradient norm-based rewards influence the teacher's ability to craft challenging yet achievable learning sequences, ultimately enhancing the student's performance. Our results show that this approach not only accelerates the learning process but also leads to improved generalization and adaptability in complex tasks. The findings underscore the potential of gradient norm signals in creating more efficient and robust ACL systems, opening new avenues for research in curriculum learning and reinforcement learning.

\noindent This study\footnote{\url{https://github.com/riensou/automatic_curriculum_learning}} explores an approach to Automatic Curriculum Learning (ACL) in reinforcement learning (RL), focusing on the integration of gradient norm reward signals. Traditional ACL methods primarily rely on predefined metrics that may not adequately capture the intricacies of learning dynamics. Our work proposes a method where the `teacher' algorithm adapts the curriculum based on the gradient norm information from the `student' model, hypothesizing that these signals offer a more refined insight into the student's learning progression. \\

\noindent We developed a framework (Teacher-Student ACL) where the teacher dynamically adjusts the learning curriculum, guided by the gradient norms of the student's model. This approach allows for a more responsive and tailored learning experience. The experimental setup utilized the PointMaze, AntMaze, and AdroitHandRelocate environments. These diverse settings provided a robust platform to assess the versatility and effectiveness of our proposed method. \\

\noindent Our experiments showed that incorporating gradient norm-based rewards significantly impacts the learning curve of the student model. In PointMaze, we observed a marked acceleration in learning, with the student models achieving proficiency more rapidly with the use of a teacher trained on gradient reward signals compared to without any teacher. In the AdroitHandRelocate environment, the student models demonstrated not only faster learning but also improved evaluation return scores. These results suggest that gradient norm signals offer a more nuanced understanding of learning progression, enabling the teacher to devise more effective and challenging curriculums. \\

\noindent The implications of our findings are substantial for the future of curriculum learning in RL. The ability of the gradient norm signals to provide detailed insights into the learning process opens new possibilities for more efficient training of RL models. This could lead to advancements in various applications, from autonomous systems to complex decision-making tasks. Furthermore, our approach underscores the importance of adaptive and dynamic curriculum strategies in enhancing the learning capabilities of AI systems. \\

\noindent Our study establishes the potential of gradient norm reward signals. Future research could explore the integration of this approach in other learning paradigms, further expanding its applicability and impact. Additionally, comparisons to other ACL methods that use other reward signals could lead to further understanding of the efficacy of gradient reward signals.

\noindent 

\end{abstract}

% \vspace{10pt}
\newpage

\section{Introduction}

Curriculum learning has emerged as a significant paradigm in artificial intelligence, offering a structured and efficient approach to training machine learning models. This concept, drawing inspiration from the way humans learn, involves progressively training models on tasks of increasing complexity. Curriculum learning stands on the premise that learning is more effective when it follows a certain order of increasing difficulty. Just as a human student learns better when concepts are introduced from simple to complex, machine learning models to benefit from a similar approach. This method contrasts with the traditional random presentation of data, offering a more guided and focused learning trajectory. \cite{soviany2022curriculum} \cite{9392296} \\

In domains such as natural language processing and computer vision, curriculum learning has shown to improve model performance, especially in dealing with complex tasks. By starting with simpler, easier-to-learn examples and gradually introducing more challenging data, models can develop a robust understanding of the underlying patterns, leading to better generalization and performance. \\

Recent developments in RL focus on dynamic curriculum strategies that adjust based on the agent's performance, optimizing the training process and potentially leading to breakthroughs in more sophisticated and adaptive RL systems. Our work attempts to explore the intricacies and benefits of applying curriculum learning principles to reinforcement learning environments. Our desired benefit is that by starting with less complex tasks, agents can quickly grasp basic skills, which can be incrementally built upon for increasingly complex tasks. In particular, we are most interested in being able to automatically generate “good curriculums.” While curriculum learning is met with widespread success, there is no clear evaluation method on the effectiveness of a given curriculum. \cite{electronics12071676}  \cite{pmlr-v139-romac21a}  

\section{Related Works}

\subsection{Curriculum Learning for Language Modeling}

In the domain of language modeling, curriculum learning has been effectively employed to enhance model training and performance. The application of curriculum learning applied to Neural machine translation achieved notable improvements in translation quality. More recently, this approach has been adapted to the pre-training phases of transformer models, enhancing their efficiency in handling complex language tasks. \cite{soviany2022curriculum}

\subsection{Curriculum Learning for Vision}

In the field of computer vision, curriculum learning has also shown promising results in fields such as object recognition, where models were gradually exposed to increasingly complex objects and scenes. Incremental learning of features and its integration with convolutional neural networks, led to more effective training and better generalization in tasks like image classification and object detection. \cite{soviany2022curriculum}

\subsection{Curriculum Learning for Reinforcement Learning}

In reinforcement learning, curriculum learning has been employed to structure the learning process of agents, beginning with simpler and progressively moving to more complex environments. This step-by-step approach has not only improved learning efficiency but also enhanced policy generalization in agents, as evidenced in various studies. Its applications range from sequencing tasks for robotic learning to adjusting difficulty levels in game environments, contributing to more robust and adaptable learning processes. Recent developments in RL focus on dynamic curriculum strategies that adjust based on the agent's performance, optimizing the training process and potentially leading to breakthroughs in more sophisticated and adaptive RL systems. \cite{electronics12071676}
\\
\\
The work of curriculum generation is an evolving work in RL as well. One way of curriculum generation is a regret-based curriculum, adapting the training distribution over the parameters of an environment, constantly producing levels at the frontier of an agent’s capabilities, resulting in curricula that start simple but become increasingly complex.
\section{Formal Background}

\subsection{Automatic Curriculum Learning}

In ACL, there is a student and a teacher. The student can be thought of as a regular DRL agent, while the teacher can be thought of as a meta-learner attempting to maximize the sample efficiency of the student. \\

Consider the Markov Decision Process (MDP) of the student, represented as $(\mathcal{S}, \mathcal{A}, \mathcal{P}, \mathcal{R}, \rho_0)$. Here $\mathcal{S}$ is the state space, $\mathcal{A}$ is the action space, $\mathcal{P}$ is the transition function, $\mathcal{R}$ is the reward function, and $\rho_0$ is the distribution of initial states. In standard DRL algorithms, the goal is to learn a policy $\pi_\theta$. \\

Consider the following ACL problem: learn a task-selection function $\pi_\phi:\mathcal{H}\to\mathcal{T}$, where $\mathcal{H}$ contains information about the student's progress and $\mathcal{T}$ is space of tasks assignable by the teacher \cite{portelas2020automatic}. In this paper, we use the student's rewards from previous assignments as $\mathcal{H}$ and initial states, $\rho_0$, as tasks assignable by the teacher. Thus, this problem can be understood as another MDP for the teacher. States in this MDP are represented by initial state and reward pairs, $(\rho_0, r)$. Actions in this MDP are given by initial state assignments to the student, $\rho_0$. Finally, the rewards are given by some measure of learning progress by the student, denoted by $\nabla\theta$ for now.

\subsection{Teacher-Student ACL Algorithm}

\begin{figure}[h]  
  \centering  
  \includegraphics[width=0.6\textwidth]{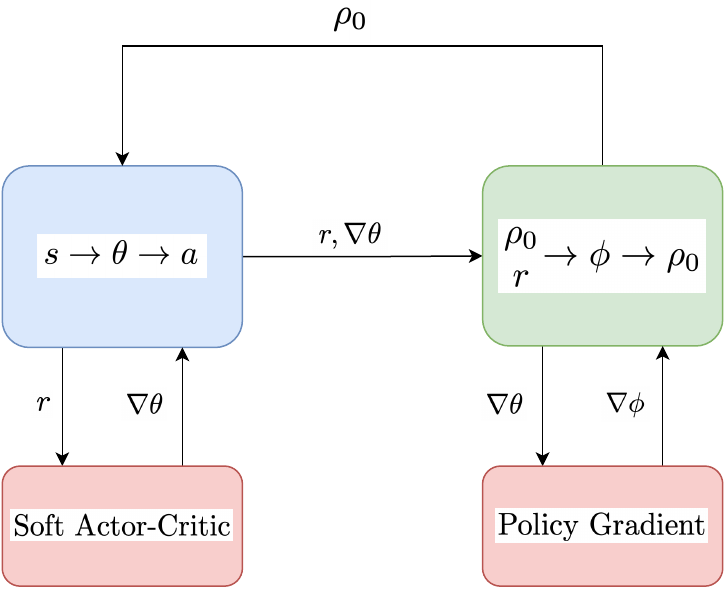}
  \caption{The Teacher-Student ACL process}
  \label{fig:1} 
\end{figure}

Now that the student and teacher MDPs have been formalized, consider how one might go about training the teacher agent. Let $A_S$ and $A_T$ be DRL algorithms, where $A_T$ is an on-policy DRL algorithm. Use the teacher to generate the student dataset, $\mathcal{D}_S$, and facilitate student learning by using $A_S$. The outcome of learning from $A_S$ can then be used to construct the teacher dataset, $\mathcal{D}_T$. Next, use $A_T$ to facilitate teacher learning. Repeat this process of student learning followed by teacher learning until the student has achieved some threshold set beforehand. See Algorithm \ref{algorithm:acl} for a more concrete explanation of this process, or Figure \ref{fig:1} for a visualization. \\

\begin{algorithm}
\caption{Teacher-Student ACL}\label{algorithm:acl}
\begin{algorithmic}[1]
\State \textbf{Input:} DRL Algorithms $A_S$, $A_T$\;
\State \textbf{Output:} $\pi_\theta$, $\pi_\phi$\;
\State Repeat:
\State \quad Repeat $k$ times:
\State \quad \quad $\mathcal{D}_S=\pi_\phi(\mathcal{H})$
\State \quad \quad $\nabla\theta= A_S(\pi_\theta,\mathcal{D}_S)$
\State \quad \quad $\theta\leftarrow\theta-\alpha_S\nabla\theta$
\State \quad $\mathcal{D}_T= \left\{(\rho_0,r)_i,(\rho_0)_i,(\nabla\theta)_i\right\}_{i=1}^k$
\State \quad $\nabla\phi=A_T(\pi_\phi,\mathcal{D}_T)$
\State \quad $\phi\leftarrow\phi-\alpha_T\nabla\phi$
\State Return $\pi_\theta$, $\pi_\phi$
\end{algorithmic}
\end{algorithm}

The hope is that the teacher will learn which environments increase the rate of student learning, thus reducing the number of samples needed otherwise. This is distinct from learning an efficient exploration scheme however, as the teacher controls starting states rather than actions. \\

\section{Methodology}

\subsection{Implementing Teacher-Student ACL}

Many design choices are left to made from Algorithm \ref{algorithm:acl} before actually implementing anything. The two DRL Algorithms $A_S$ and $A_T$, as well as the metric of learning progress, $\nabla\theta$ are all important decisions that need to be made. For $A_S$ we used soft actor-critic and for $A_T$ we used policy gradient. For learning progress, many metrics have been used in the past such as student loss, model complexity increases, and others \cite{graves2017automated}. We focused on two novel measures of learning progress, both based on the student gradients encountered during training on the teacher's assignments. These metrics of learning progress are as follows:
\begin{enumerate}
    \item Average gradient size during episode:
    $$\frac{1}{T}\sum_{t=1}^T\|\left(\nabla\theta\right)_t\|_2$$
    \item Size of overall episode gradient:
    $$\frac{1}{T}\left\|\sum_{t=1}^T\left(\nabla\theta\right)_t\right\|_2$$
\end{enumerate}

The first metric, the average gradient size during episode, accounts for how much the student was able to learn at each step starting from the initial state given by the teacher. This may be better than the alternative as it still values steps that slightly counteract each other. The second metric, the size of the overall episode gradient, accounts for the total change that comes from the episode starting from the initial state given by the teacher. 

\subsection{Training Data}

\begin{figure}[h]
    \centering
    \begin{subfigure}[b]{0.3\textwidth}
        \includegraphics[width=\textwidth]{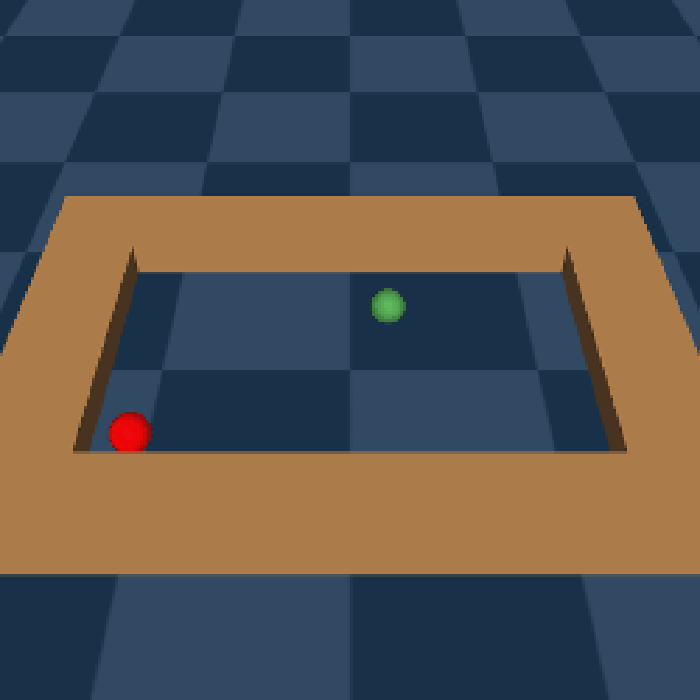}
        \caption{PointMaze\_OpenDense-v3}
        \label{fig:sub11}
    \end{subfigure}
    \begin{subfigure}[b]{0.3\textwidth}
        \includegraphics[width=\textwidth]{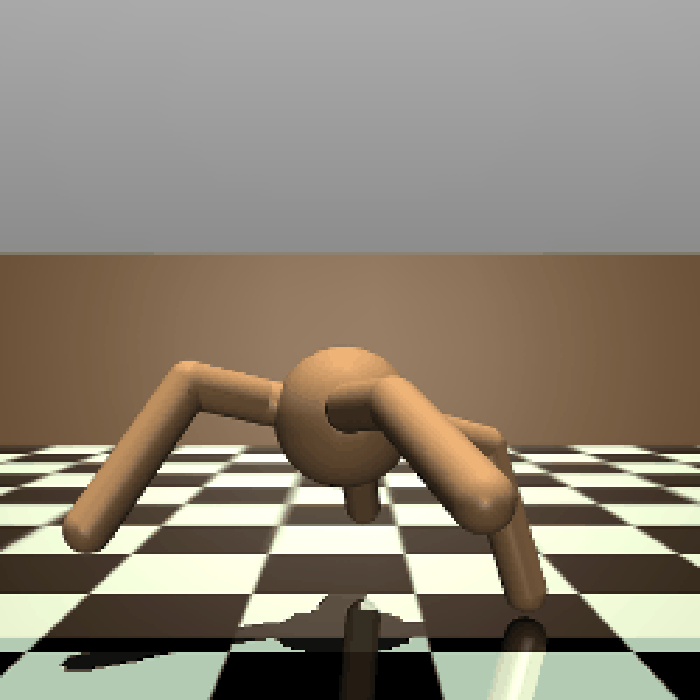}
        \caption{AntMaze\_OpenDense-v4}
        \label{fig:sub21}
    \end{subfigure}
    \begin{subfigure}[b]{0.3\textwidth}
        \includegraphics[width=\textwidth]{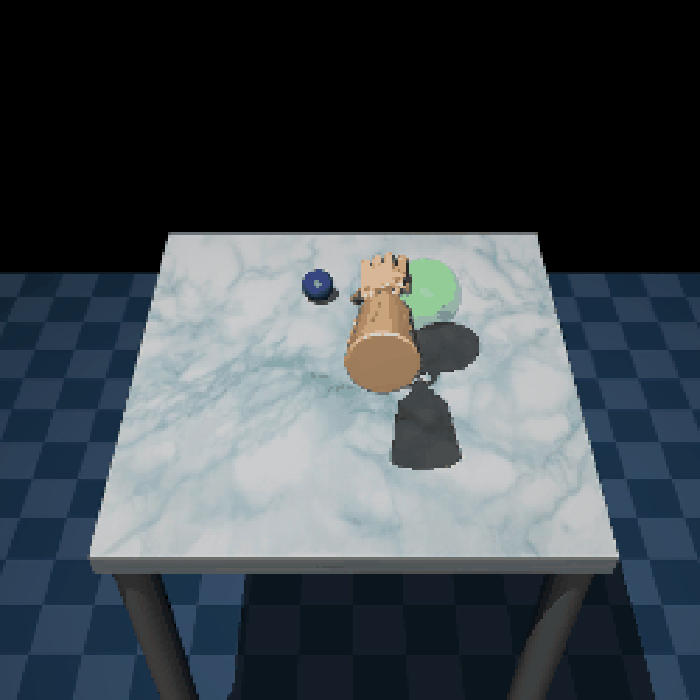}
        \caption{AdroitHandRelocate-v1}
        \label{fig:sub32}
    \end{subfigure}
    \caption{The three different gym environments used in this paper}
    \label{fig:2}
\end{figure}

Our research utilizes OpenAI gym environments \cite{brockman2016openai}, particularly the PointMaze, AntMaze, and AdroitHandRelocate environments \cite{gymnasium_robotics2023github}, as seen in Figure \ref{fig:2}. The PointMaze and AntMaze environments are designed for testing reinforcement learning algorithms in navigation and pathfinding tasks. The AdroitHandRelocate environment is a more complicated environment that is designed to test reinforcement learning algorithms in spatial awareness and decision-making. Additionally, the potential starting states of the AdroitHandRelocate environment are much higher-dimensional than the mazes, giving the teacher more control over the student's curriculum.

\subsection{Model Architecture / Configurations}
All our experiments were performed with the following model configurations. These can be found in more depth in the corresponding yaml files. 

\begin{itemize}
  \item \textbf{Total Steps}: 1,000,000
  \item \textbf{Batch Size}: 256
  \item \textbf{Hidden Layer Size}: 256
  \item \textbf{Number of Layers}: 3
  \item \textbf{Discount Factor}: 0.99
  \item \textbf{Temperature}: 0.05
\end{itemize}

\subsection{Smoothing}

Each experiment ran for at least $750,000$ steps, and many values were logged frequently. Thus, many graphs look very chaotic and are hard to interpret on their own. Thus, the following smoothing technique was used: $\Tilde{y}_i=\lambda\Tilde{y}_{i-1}+(1-\lambda)y_i$. Larger $\lambda$ values (closer to $1$) were used whenever the data was logged more frequently. The raw, unsmoothed graphs can all be found in the appendix.

\section{Findings}
To evaluate the effectiveness of teaching, we employed several experiments comparing the students with (1) a teacher with average gradient per time-step as a reward, (2) a teacher with overall episode gradient as a reward, and (3) no teacher. Our findings suggest that curriculum learning with a teacher using gradients as a reward signal is effective. This effect is pronounced after label smoothing, as described in the above section. For most experiments, either one or both teaching methods were shown to outperform the no-teacher baseline, either converging to the peak faster, or having a bigger peak overall. Subsections below discuss these results in more depth. Dissecting the curriculum learned by the teacher showed a clear improvement in the teacher in generating harder tasks over time, which we approximate by the average goal distance that the teacher's generated curriculum provides. 

\subsection{PointMaze Performance}

\begin{figure}[h]
    \centering
    \begin{subfigure}[b]{0.45\textwidth}
        \includegraphics[width=\textwidth]{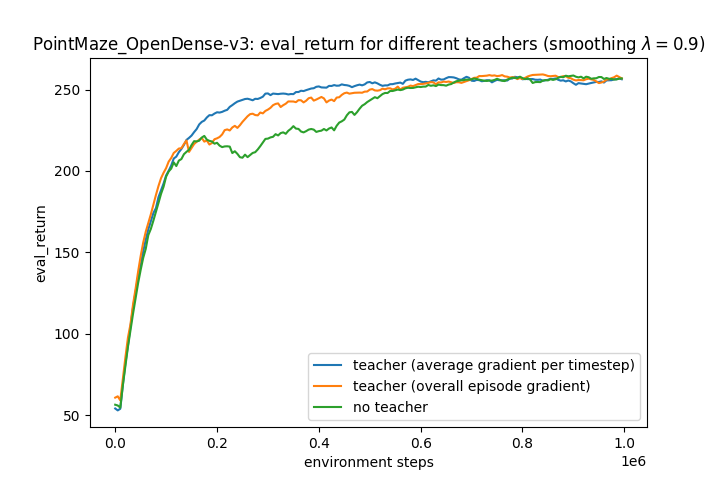}
        \caption{Eval Return on PointMaze environment}
        \label{fig:5b}
    \end{subfigure}
    \begin{subfigure}[b]{0.45\textwidth}
        \includegraphics[width=\textwidth]{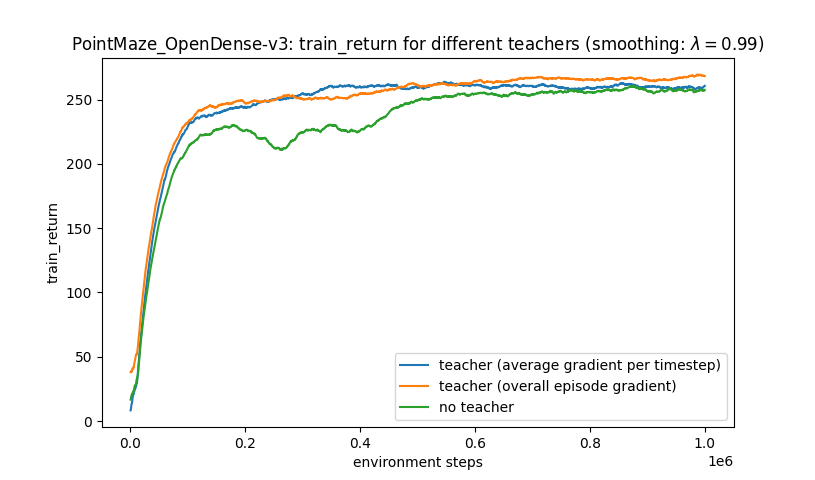}
        \caption{Train Return on PointMaze environment}
        \label{fig:6b}
    \end{subfigure}
    \caption{PointMaze}
    \label{fig:pointmaze_exps}
\end{figure}

As shown in Figure \ref{fig:pointmaze_exps}, the train return and eval return converge faster with a teacher than compared to without a teacher. The different teachers tested are the teachers trained with different rewards. \\

While each agent is able to converge, it does seem that the agent with the teacher using average gradient per time-step is slightly faster than the teacher using overall episode gradient. This is more clear from the eval return, while the train return they seem to be indistinguishable. 

\subsection{AntMaze Performance}

\begin{figure}[h]
    \centering
    \includegraphics[width=0.7\textwidth]{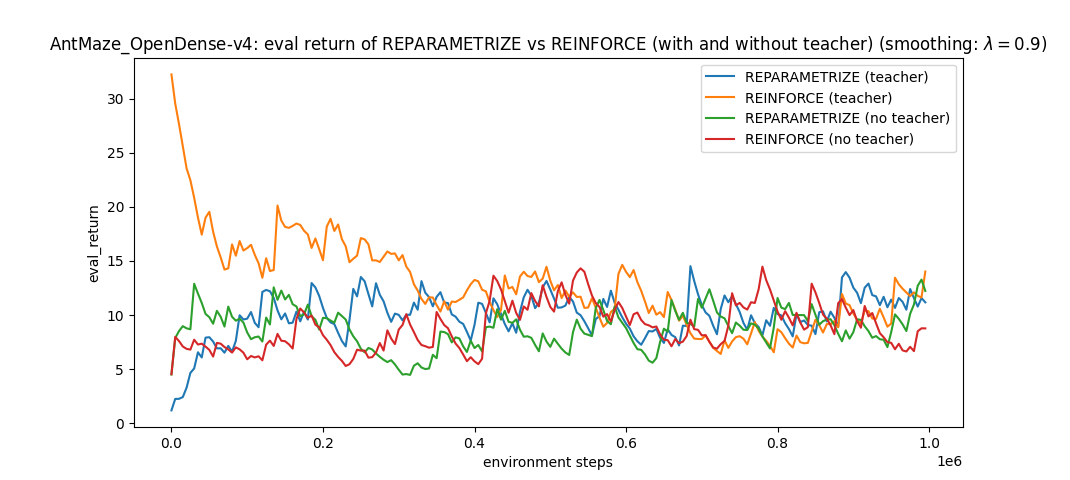}
    \caption{Teacher vs No Teacher on AntMaze environment}
    \label{fig:1b}
\end{figure}

We were unable to get a significant outcome employing teaching on the AntMaze environment. As exhibited by Figure \ref{fig:1b}, the agents with a teacher and no teacher all performed similarly after a million environment steps. Note that slight variations of $A_S$ were also tested (reparametrize, reinforce), but no significant learning was able to happen regardless. This suggests that the AntMaze environment is either too challenging for the capacity of the models used or the learning algorithms employed. 

\subsection{AdroitHandRelocate Performance}

\begin{figure}[h]
    \centering
    \begin{subfigure}[b]{0.45\textwidth}
        \includegraphics[width=\textwidth]{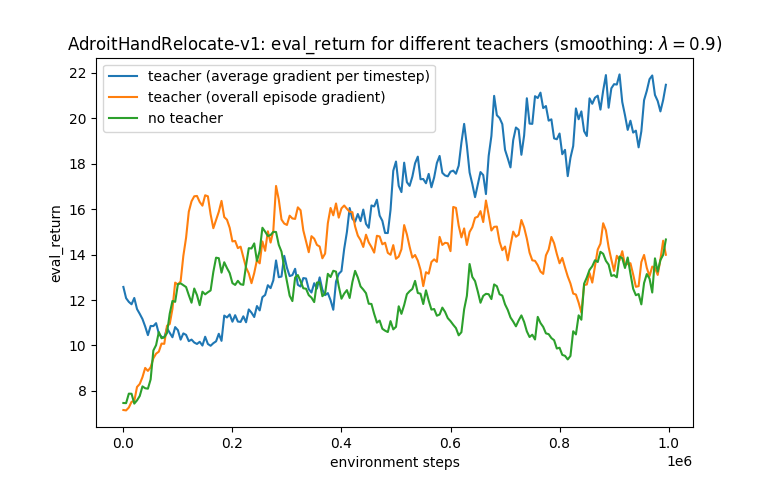}
        \caption{Eval Return on AdroitHandRelocate environment}
        \label{fig:2b}
    \end{subfigure}
    \begin{subfigure}[b]{0.45\textwidth}
        \includegraphics[width=\textwidth]{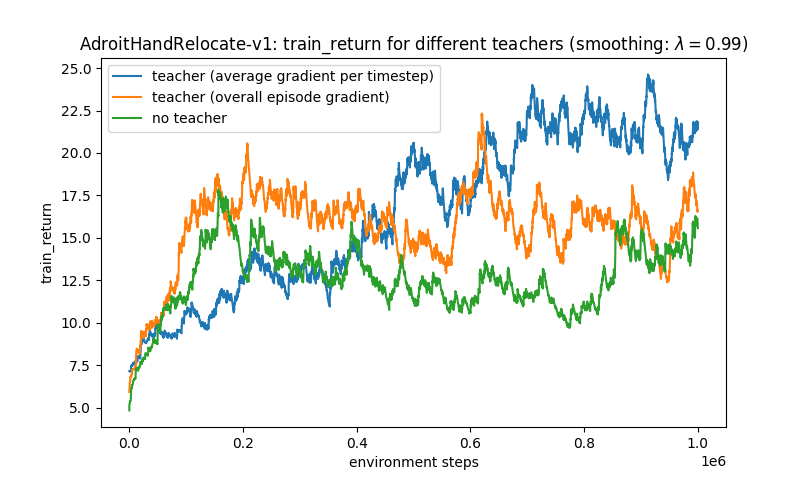}
        \caption{Train Return on AdroitHandRelocate environment}
        \label{fig:3b}
    \end{subfigure}
    \caption{AdroitHandRelocate}
    \label{fig:adroithandsrelocate}
\end{figure}

In the AdroitHandRelocate environment, the teacher using average gradient per time-step outperformed both the agent with the teacher using overall episode gradient and no teacher, as shown in Figure \ref{fig:adroithandsrelocate}. The overall episode gradient over-performed no teacher throughout the environment steps, until converging towards a similar return near the end. 

\subsection{Teacher Effect on Gradients}

\begin{figure}[h]
    \centering
    \begin{subfigure}[b]{0.45\textwidth}
        \includegraphics[width=\textwidth]{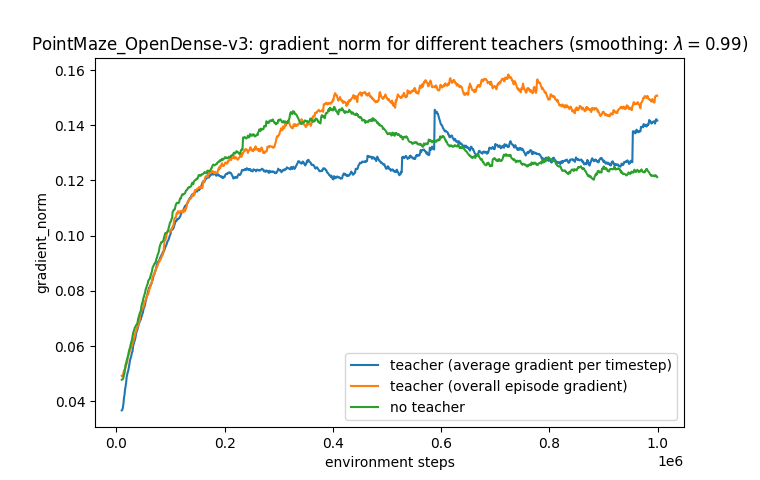}
        \caption{Gradient norms on PointMaze environment}
        \label{fig:7b}
    \end{subfigure}
    \begin{subfigure}[b]{0.45\textwidth}
        \includegraphics[width=\textwidth]{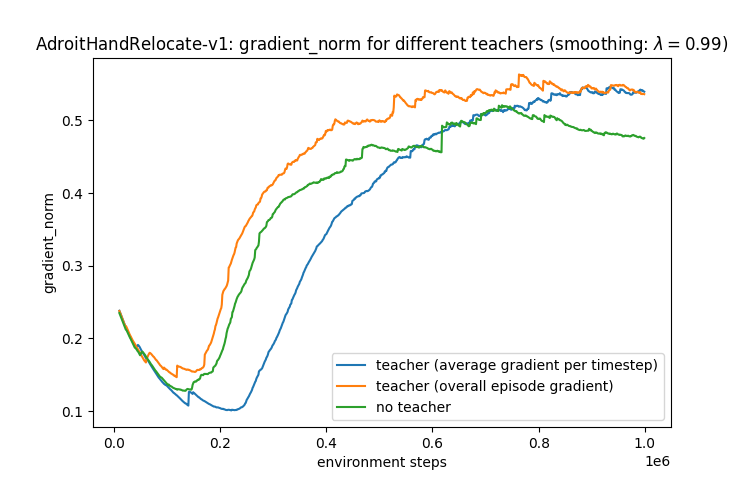}
        \caption{Gradient norms on AdroitHandRelocate environment}
        \label{fig:4b}
    \end{subfigure}
    \caption{Gradient norms over time for different teachers}
    \label{fig:gradnorms}
\end{figure}

As shown in Figure \ref{fig:gradnorms}, the student's gradient norms varied over time in ways that depended on the type of teacher present. In both the PointMaze and AdroitHandRelocate environments, the agent with the teacher with average gradient per time-step rewards started with the lowest gradients, but eventually caught up to the agent with the teacher with overall episode gradient rewards. \\

This is surprising, as the agent with the teacher with average gradient per time-step as rewards performed the best in terms of eval return. Thus, this seems to indicate that while the gradient norms are smaller, they are more precise and in the correct direction. 

\subsection{Visualizing the Learned Curriculum}

\begin{figure}[h]
    \centering
    \includegraphics[width=0.7\textwidth]{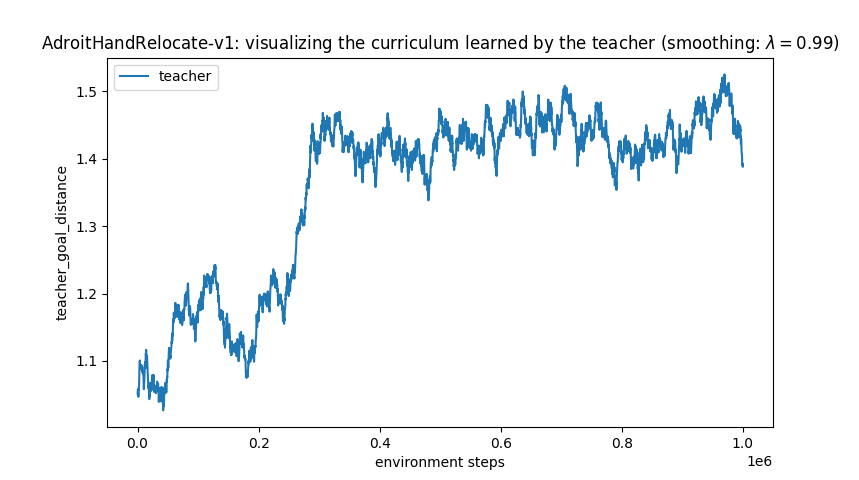}
    \caption{Curriculum learned by teacher with respect to goal distance}
    \label{fig:13b}
\end{figure}

In order to visualize the learned curriculum of the teacher, the distance of the ball to the target was tracked in the AdroitHandRelocate environment. Intuitively, one would expect an easier curriculum to put the ball further and further away as the agent gets better at moving it, and this is what can be seen in Figure \ref{fig:13b}. 

This aligns with the intuitive reasoning that harder tasks are associated with farther goals, meaning that the teacher correctly learns to generate more difficult curriculum over time, to improve the performance of the agent. \\

Additionally, it should be noted that the teacher had control over other things such as initial hand position and velocity, so the teacher did more than just move the ball away. 

\section{Conclusion}

% Through the Teacher-Student ACL algorithm, we have successfully created an instance of using the gradients of the student to train a teacher. The teacher succeeded in boosting the performance of the student in terms of both sample efficiency and peak return, compared to the student's learning algorithm without the teacher. \\

% Furthermore, inspecting the curriculum learned by the teacher shows that the teacher learned how to generate curriculums that make intuitive sense, as shown by the the increasing goal distance that a teacher generates, aligning with the performance of the agent. \\

% This work opens up the possibility of testing different reward signals based on gradient norms. For example, it may be interesting to focus on the gradients of different layers of the agent network. Maybe using the final layer's gradients could be an interesting reward signal for the teacher. Also, having the teacher only have the ability to control initial states seems limiting, and it may be interesting to use gradients as reward signals in contexts where the teacher has more control. This would give the teacher more data to train on, and thus might further improve the performance. 

Through the implementation of our Teacher-Student ACL algorithm, we have created a novel instance of using student model gradients to effectively train a teacher model. The teacher model, guided by the feedback from the student's gradients, succeeded in significantly enhancing the student's learning process. This is evident in the marked improvements in sample efficiency and peak return, surpassing the outcomes when the student model operates without the influence of the teacher. \\

A deeper examination of the curriculum devised by the teacher model reveals a sophisticated understanding of the learning trajectory. The progression in goal distances set by the teacher aligns remarkably with the improving performance of the agent, underscoring the teacher model's capacity to intuitively and dynamically adjust the learning path to optimize outcomes. This insight into curriculum development is a testament to the effectiveness of our gradient-norm-based approach in shaping a more adept and responsive teaching strategy. \\

Looking ahead, this study opens several exciting avenues for future exploration. One particularly promising direction is the investigation of different reward signals based on gradient norms. Focusing on the gradients of specific layers within the agent's network, such as the final layer, might yield intriguing insights and further enhance the teacher's capability to fine-tune the learning process. Moreover, the current restriction to initial state control for the teacher can be seen as a starting point rather than a limitation. Expanding the teacher's control to encompass more aspects of the learning environment could provide a richer dataset for the teacher to learn from. Such an expansion of control parameters could lead to a more comprehensive and effective training regime, potentially unlocking new levels of performance and efficiency in student models. \\

In conclusion, our research presents a meaningful advancement in the use of gradient norms for ACL. This approach not only improves the performance of student models in reinforcement learning but also opens up new possibilities for refining training processes. Future investigations could explore different gradient-based reward signals and expand the teacher's control within the learning environment, potentially leading to more nuanced and effective training methods. This study contributes to the ongoing evolution of AI training strategies, suggesting new paths for exploration and development.

% I don't really know what this is trying to say...
% Additionally, we believe that this work opens up the possibility of evaluating new heuristics on the difficulty of a task. Visualizing the curriculum learned by the teacher, with respect to goal distance, showed us a clear relationship between goal distance and maze difficulty. Inspecting other variables may reveal more heuristics on evaluating environment difficulty. As opposed to the current literature which evaluates environment difficulty based on agent performance, we believe we have demonstrated a potential for environment difficulty to be estimated by teacher inspection. 

\newpage

\bibliographystyle{plain} % We choose the "plain" reference style
\bibliography{refs} % Entries are in the refs.bib file

\begin{thebibliography}{1}

\bibitem{brockman2016openai}
Greg Brockman, Vicki Cheung, Ludwig Pettersson, Jonas Schneider, John Schulman, Jie Tang, and Wojciech Zaremba.
\newblock Openai gym, 2016.

\bibitem{gymnasium_robotics2023github}
Rodrigo de~Lazcano, Kallinteris Andreas, Jun~Jet Tai, Seungjae~Ryan Lee, and Jordan Terry.
\newblock Gymnasium robotics, 2023.

\bibitem{graves2017automated}
Alex Graves, Marc~G. Bellemare, Jacob Menick, Remi Munos, and Koray Kavukcuoglu.
\newblock Automated curriculum learning for neural networks, 2017.

\bibitem{electronics12071676}
Fengchun Liu, Tong Zhang, Chunying Zhang, Lu~Liu, Liya Wang, and Bin Liu.
\newblock A review of the evaluation system for curriculum learning.
\newblock {\em Electronics}, 12(7), 2023.

\bibitem{portelas2020automatic}
Rémy Portelas, Cédric Colas, Lilian Weng, Katja Hofmann, and Pierre-Yves Oudeyer.
\newblock Automatic curriculum learning for deep rl: A short survey, 2020.

\bibitem{pmlr-v139-romac21a}
Cl{\'e}ment Romac, R{\'e}my Portelas, Katja Hofmann, and Pierre-Yves Oudeyer.
\newblock Teachmyagent: a benchmark for automatic curriculum learning in deep rl.
\newblock In Marina Meila and Tong Zhang, editors, {\em Proceedings of the 38th International Conference on Machine Learning}, volume 139 of {\em Proceedings of Machine Learning Research}, pages 9052--9063. PMLR, 18--24 Jul 2021.

\bibitem{soviany2022curriculum}
Petru Soviany, Radu~Tudor Ionescu, Paolo Rota, and Nicu Sebe.
\newblock Curriculum learning: A survey, 2022.

\bibitem{9392296}
Xin Wang, Yudong Chen, and Wenwu Zhu.
\newblock A survey on curriculum learning.
\newblock {\em IEEE Transactions on Pattern Analysis and Machine Intelligence}, 44(9):4555--4576, 2022.

\end{thebibliography}

\section*{Appendix}

\subsection*{A: Teacher Updates and Buffer Clearing}

The following are two hyperparameter settings that were studied that seemed to not make too much of an impact, but are nonetheless interesting:
\begin{enumerate}
    \item \textbf{number of teacher updates}: determines the number of times to perform an update to the teacher network each time it is updated
    \item \textbf{clear buffer}: toggles whether or not to clear the student's replay buffer after each time the teacher gives a new initial state 
\end{enumerate}

\begin{figure}[h]
    \centering
    \begin{subfigure}[b]{0.45\textwidth}
        \includegraphics[width=\textwidth]{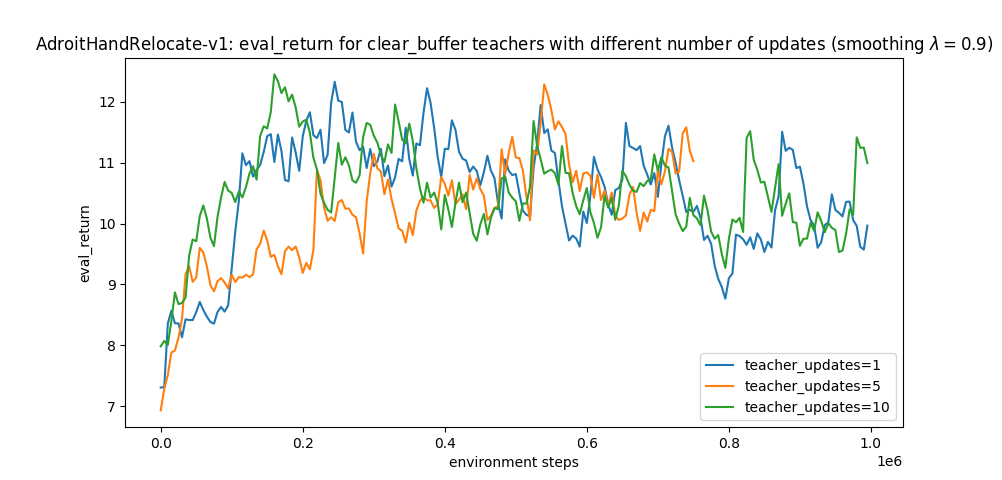}
        \caption{Eval return on AdroitHandRelocate}
        \label{fig:8b}
    \end{subfigure}
    \begin{subfigure}[b]{0.45\textwidth}
        \includegraphics[width=\textwidth]{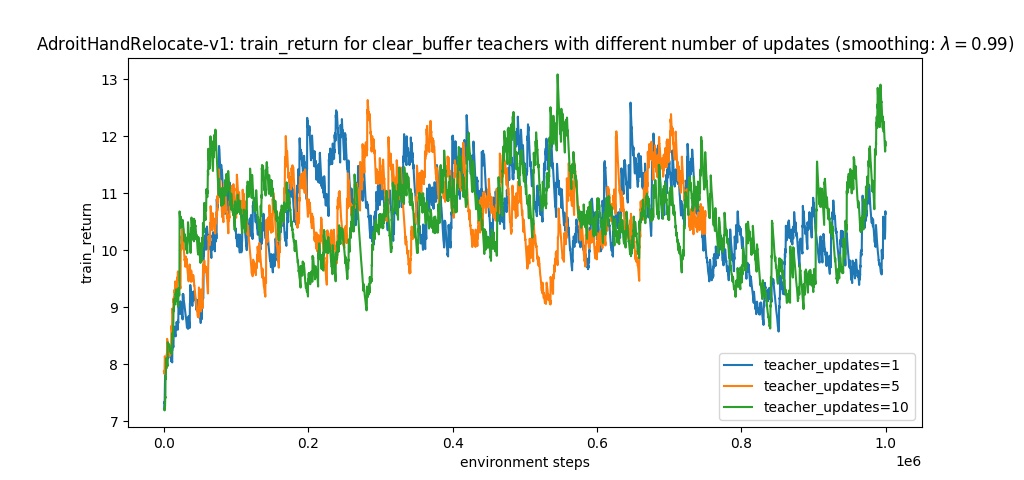}
        \caption{Train return on AdroitHandRelocate}
        \label{fig:9b}
    \end{subfigure}
    \caption{Different number of teacher updates with clear buffer}
    \label{fig:twosidebyside1}
\end{figure}

\begin{figure}[h]
    \centering
    \includegraphics[width=0.7\textwidth]{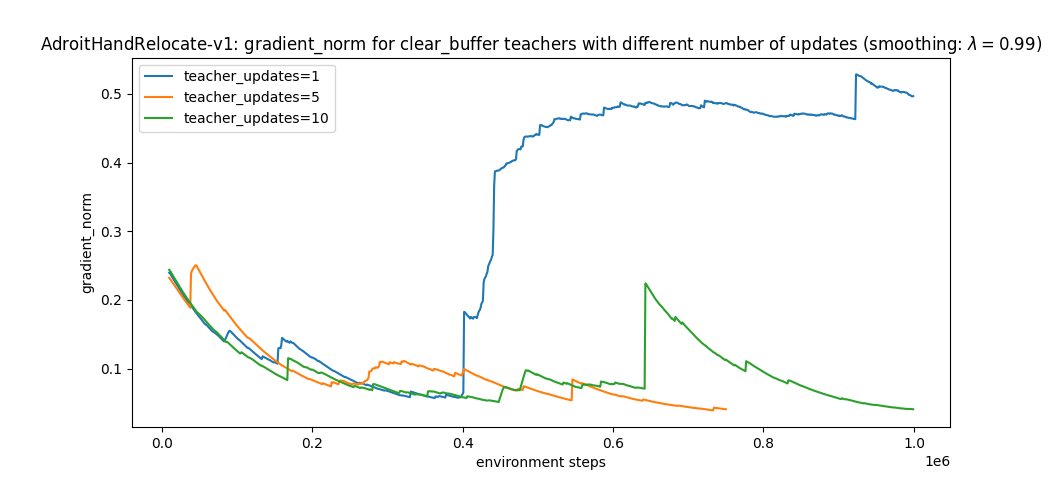}
    \caption{Gradient norm for different number of teacher updates with clear buffer}
    \label{fig:10b}
\end{figure}

\begin{figure}[h]
    \centering
    \begin{subfigure}[b]{0.45\textwidth}
        \includegraphics[width=\textwidth]{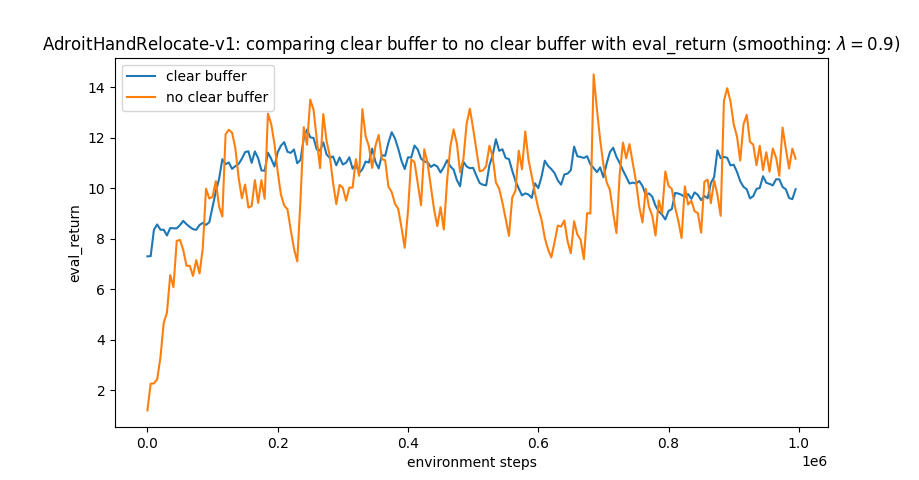}
        \caption{Eval return on AdroitHandRelocate}
        \label{fig:11b}
    \end{subfigure}
    \begin{subfigure}[b]{0.45\textwidth}
        \includegraphics[width=\textwidth]{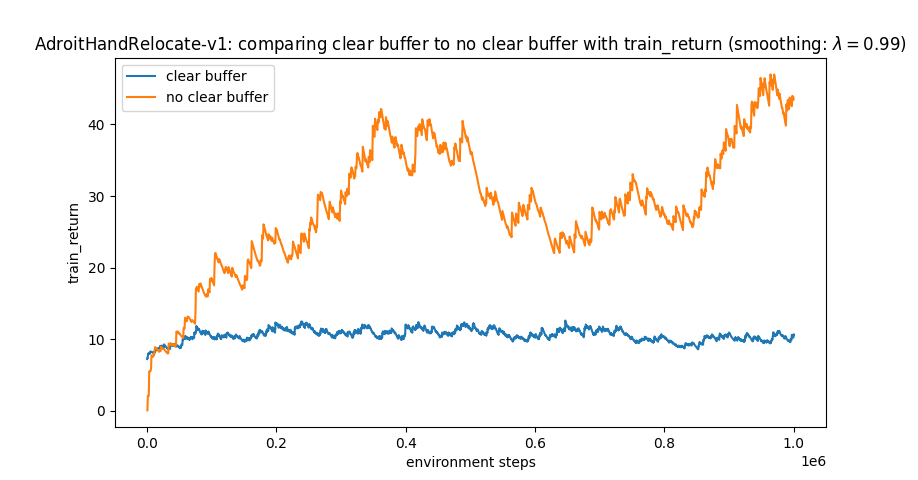}
        \caption{Train return on AdroitHandRelocate}
        \label{fig:12b}
    \end{subfigure}
    \caption{clear buffer vs no clear buffer on AdroitHandRelocate}
    \label{fig:twosidebyside4}
\end{figure}

\newpage

\subsection*{B: Unsmoothed Graphs}

\begin{figure}[h]
    \centering
    \begin{subfigure}[b]{0.3\textwidth}
        \includegraphics[width=\textwidth]{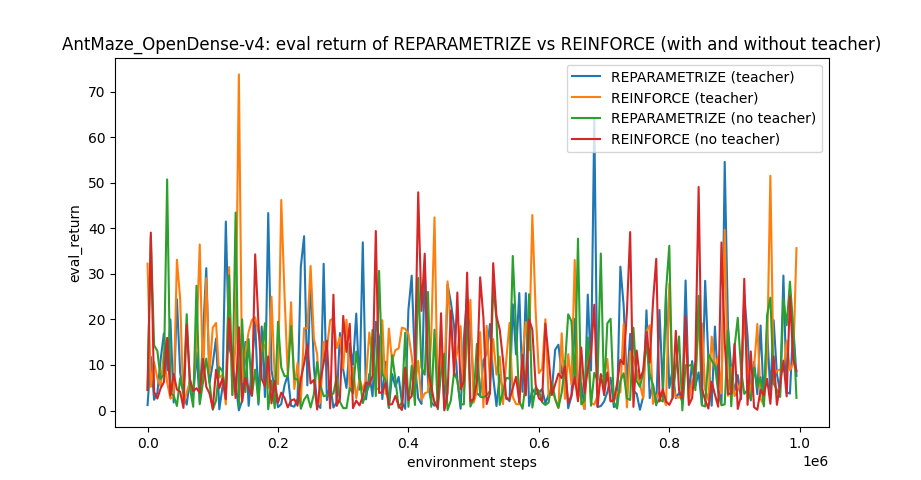}
        \caption{Unsmoothed graph of Figure \ref{fig:1b}}
        \label{fig:sub14}
    \end{subfigure}
    \begin{subfigure}[b]{0.3\textwidth}
        \includegraphics[width=\textwidth]{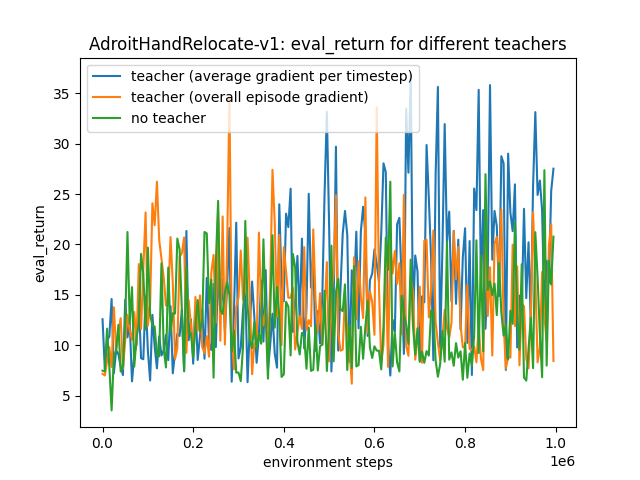}
        \caption{Unsmoothed graph of Figure \ref{fig:2b}}
        \label{fig:sub22}
    \end{subfigure}
    \begin{subfigure}[b]{0.3\textwidth}
        \includegraphics[width=\textwidth]{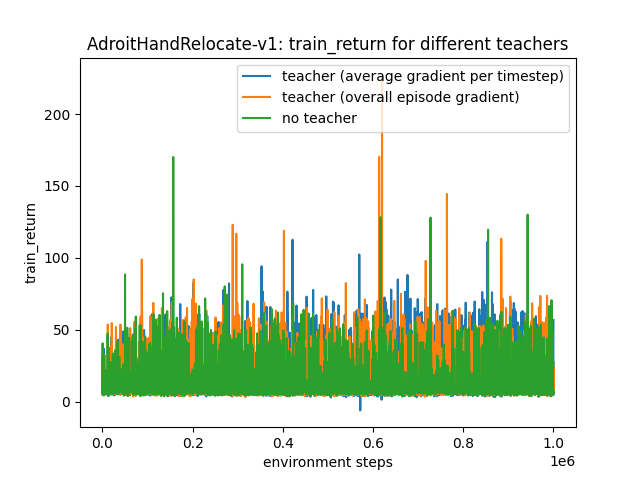}
        \caption{Unsmoothed graph of Figure \ref{fig:3b}}
        \label{fig:sub33}
    \end{subfigure}
    \caption{Unsmoothed graphs of figures}
    \label{fig:caption2}
\end{figure}

\begin{figure}[h]
    \centering
    \begin{subfigure}[b]{0.3\textwidth}
        \includegraphics[width=\textwidth]{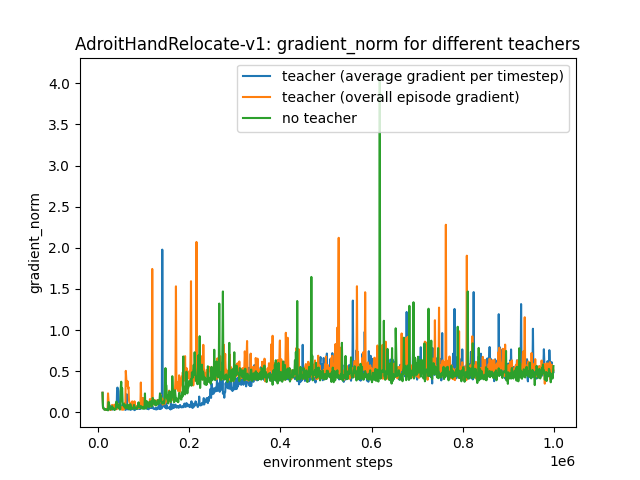}
        \caption{Unsmoothed graph of Figure \ref{fig:4b}}
        \label{fig:sub13}
    \end{subfigure}
    \begin{subfigure}[b]{0.3\textwidth}
        \includegraphics[width=\textwidth]{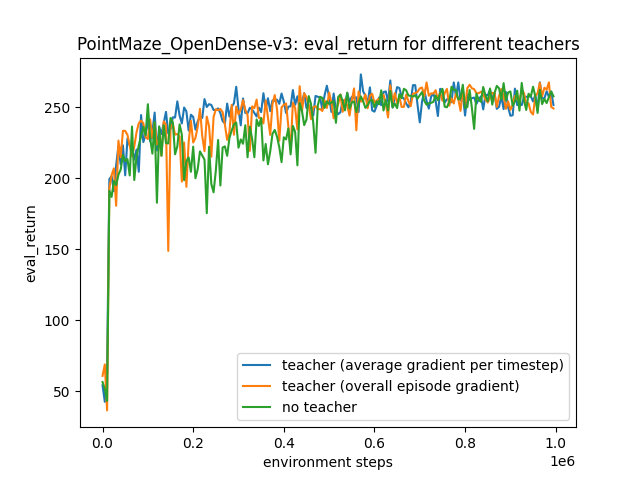}
        \caption{Unsmoothed graph of Figure \ref{fig:5b}}
        \label{fig:sub23}
    \end{subfigure}
    \begin{subfigure}[b]{0.3\textwidth}
        \includegraphics[width=\textwidth]{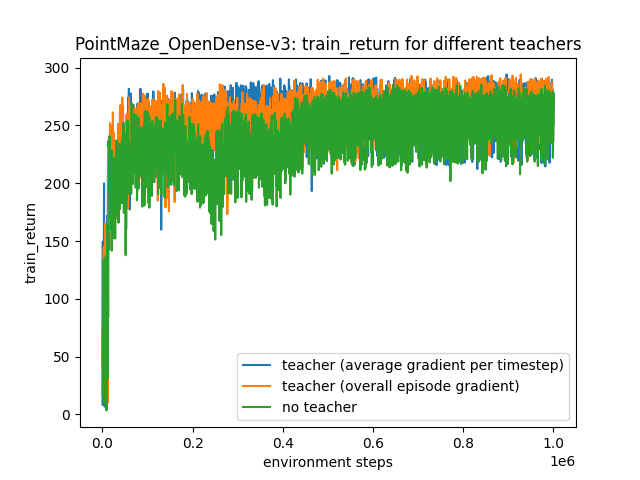}
        \caption{Unsmoothed graph of Figure \ref{fig:6b}}
        \label{fig:sub34}
    \end{subfigure}
    \caption{Unsmoothed graphs of figures}
    \label{fig:caption3}
\end{figure}

\begin{figure}[h]
    \centering
    \begin{subfigure}[b]{0.3\textwidth}
        \includegraphics[width=\textwidth]{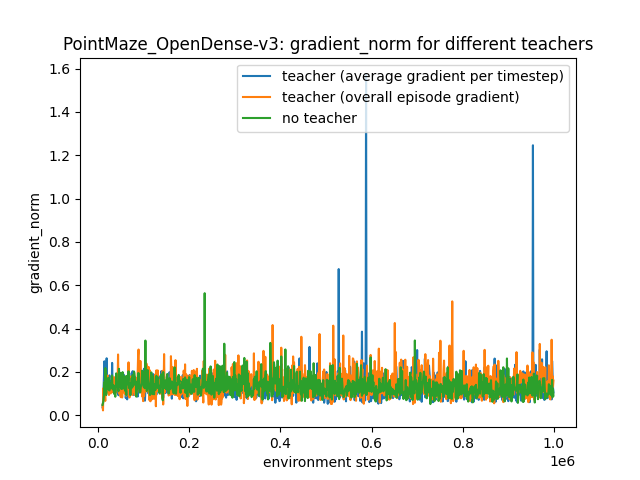}
        \caption{Unsmoothed graph of Figure \ref{fig:7b}}
        \label{fig:sub12}
    \end{subfigure}
    \begin{subfigure}[b]{0.3\textwidth}
        \includegraphics[width=\textwidth]{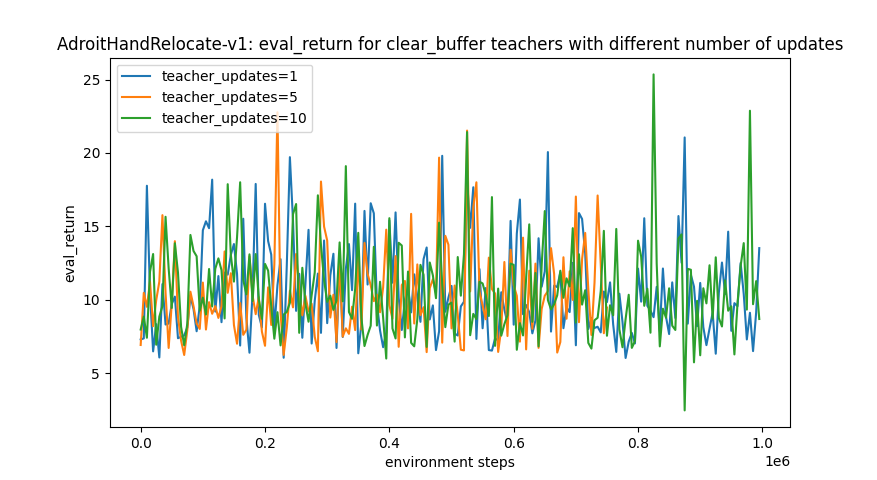}
        \caption{Unsmoothed graph of Figure \ref{fig:8b}}
        \label{fig:sub24}
    \end{subfigure}
    \begin{subfigure}[b]{0.3\textwidth}
        \includegraphics[width=\textwidth]{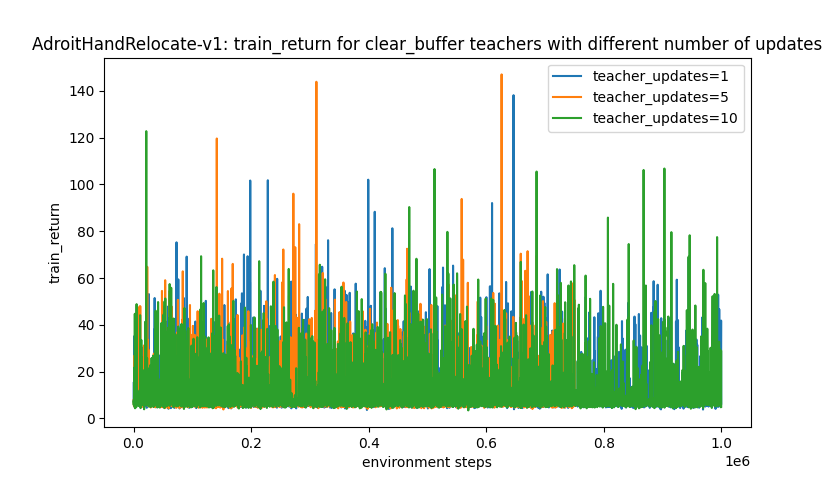}
        \caption{Unsmoothed graph of Figure \ref{fig:9b}}
        \label{fig:sub31}
    \end{subfigure}
    \caption{Unsmoothed graphs of figures}
    \label{fig:caption1}
\end{figure}

\begin{figure}
    \centering
    \begin{subfigure}[b]{0.45\textwidth}
        \includegraphics[width=\textwidth]{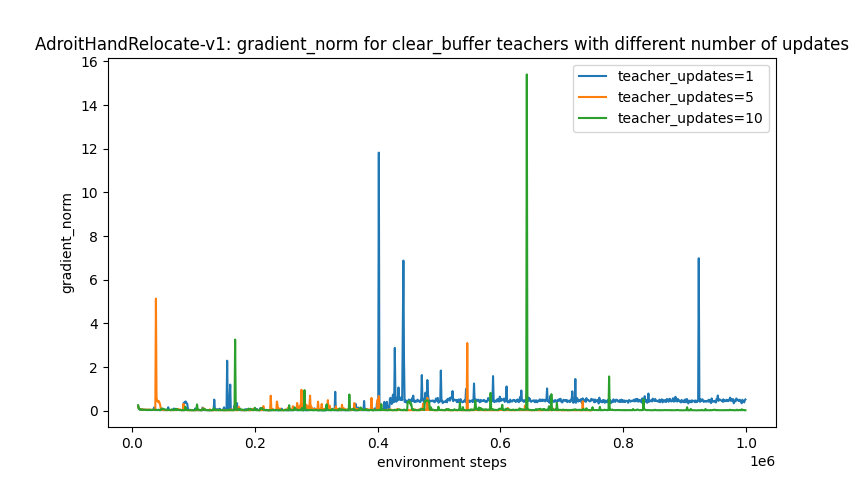}
        \caption{Unsmoothed graph of Figure \ref{fig:10b}}
        \label{fig:imagea1}
    \end{subfigure}
    \begin{subfigure}[b]{0.45\textwidth}
        \includegraphics[width=\textwidth]{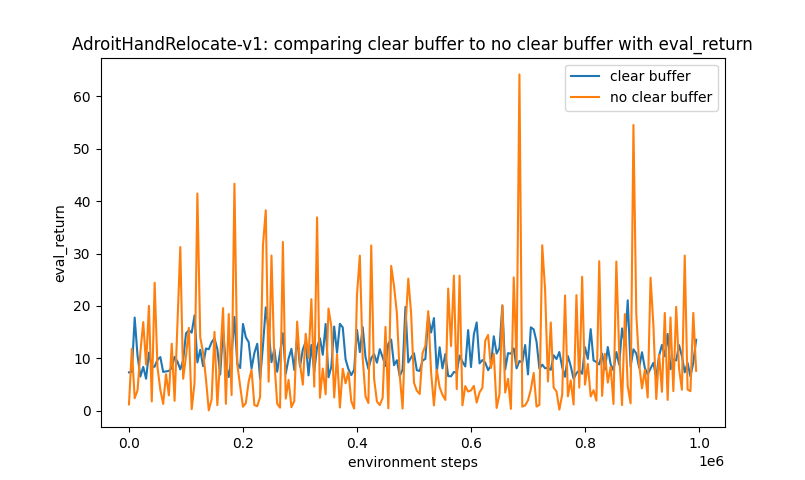}
        \caption{Unsmoothed graph of Figure \ref{fig:11b}}
        \label{fig:imageb1}
    \end{subfigure}
    \caption{Unsmoothed graphs of figures}
    \label{fig:twosidebyside3}
\end{figure}

\begin{figure}
    \centering
    \begin{subfigure}[b]{0.45\textwidth}
        \includegraphics[width=\textwidth]{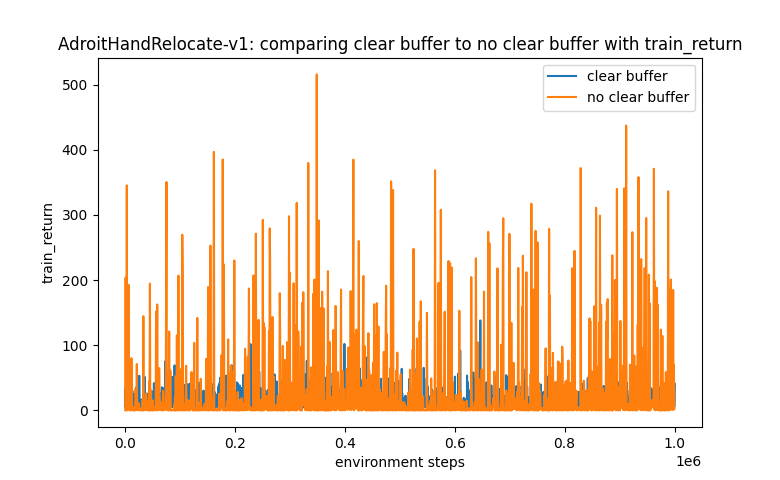}
        \caption{Unsmoothed graph of Figure \ref{fig:12b}}
        \label{fig:imagea2}
    \end{subfigure}
    \begin{subfigure}[b]{0.45\textwidth}
        \includegraphics[width=\textwidth]{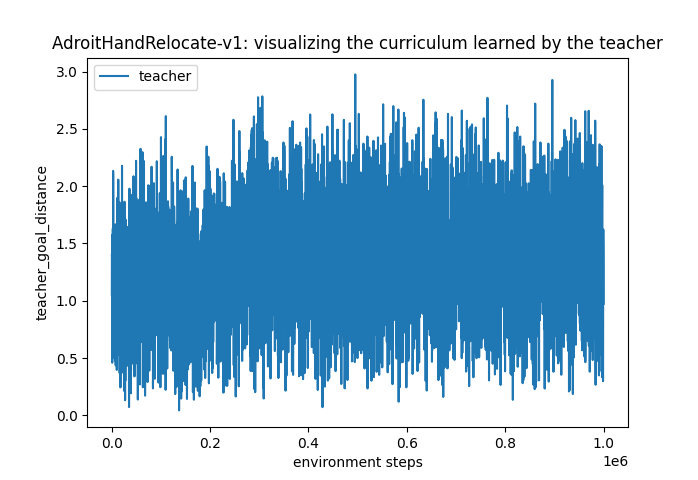}
        \caption{Unsmoothed graph of Figure \ref{fig:13b}}
        \label{fig:imageb2}
    \end{subfigure}
    \caption{Unsmoothed graphs of figures}
    \label{fig:twosidebyside2}
\end{figure}

\end{document}